\documentclass[sn-aps]{sn-jnl}

\usepackage{graphicx}%
\usepackage{multirow}%
\usepackage{amsmath,amssymb,amsfonts}%
\usepackage{amsthm}%
\usepackage{mathrsfs}%
\usepackage[title]{appendix}%
\usepackage{xcolor}%
\usepackage{textcomp}%
\usepackage{manyfoot}%
\usepackage{booktabs}%
\usepackage{algorithm}%
\usepackage{algorithmicx}%
\usepackage{algpseudocode}%
\usepackage{listings}%
\usepackage{caption}

\theoremstyle{thmstyleone}%
\theoremstyle{thmstyletwo}%

\theoremstyle{thmstylethree}%

\begin{document}

\title[Article Title]{Distillation-based fabric anomaly detection}

\author*[1,2]{\fnm{Thomine} \sur{Simon}}\email{simon.thomine@utt.fr}

\author[1]{\fnm{Snoussi} \sur{Hichem}}\email{hichem.snoussi@utt.fr}

\affil*[1]{\orgdiv{LIST3N}, \orgname{University of technology of Troyes}, \orgaddress{ \city{Troyes}, \postcode{10000}, \country{France}}}

\affil[2]{\orgname{Aquilae}, \orgaddress{ \city{Troyes}, \postcode{10000}, \country{France}}}

\abstract{Unsupervised texture anomaly detection has been a concerning topic in a vast amount of industrial processes. Patterned textures inspection, particularly in the context of fabric defect detection, is indeed a widely encountered use case. This task involves handling a diverse spectrum of colors and textile types, encompassing a wide range of fabrics. Given the extensive variability in colors, textures, and defect types, fabric defect detection poses a complex and challenging problem in the field of patterned textures inspection.
In this article, we propose a knowledge distillation-based approach tailored specifically for addressing the challenge of unsupervised anomaly detection in textures resembling fabrics. Our method aims to redefine the recently introduced reverse distillation approach, which advocates for an encoder-decoder design to mitigate classifier bias and to prevent the student from reconstructing anomalies. In this study, we present a new reverse distillation technique for the specific task of fabric defect detection. Our approach involves a meticulous design selection that strategically highlights high-level features. 
To demonstrate the capabilities of our approach both in terms of performance and inference speed, we conducted a series of experiments on multiple texture datasets, including MVTEC AD, AITEX, and TILDA, alongside conducting experiments on a dataset acquired from a textile manufacturing facility.
The main contributions of this paper are the following: a robust texture anomaly detector utilizing a reverse knowledge-distillation technique suitable for both anomaly detection and domain generalization and a novel dataset encompassing a diverse range of fabrics and defects. }


\maketitle

\section{Introduction}\label{sec1}

Fabric defect detection has been a really worrisome problem for the textile industry as the quality of the garment is a priority for the customer. Due to the absence of superior alternatives, inspection is currently carried out mainly by humans, which represents a significant labor cost and does not guarantee flawless detection due to visual fatigue and distraction. Currently, the accuracy of humans in detecting defects is around 60\% and decreases for visually difficult fabrics such as striped fabrics. Several constraints are relevant to consider when dealing with this problem: industrial processes are well oiled and cannot suffer from stoppage or slowing of any type, the variety of defects/fabric type is colossal and evolving over time with new collections, defects can be tiny and therefore require an industrial camera with high resolution. As a result, we should aim for a method that does not require defective samples in the training, while offering generalization capability and a fast inference time. 
Recently, a predominant trend in unsupervised anomaly detection methods has emerged. It involves the utilization of features extracted from pre-trained networks to learn descriptive representations of the analyzed objects or textures \cite{he_deep_2015,krizhevsky_imagenet_2017,tan_efficientnet_2020,yu_fastflow_2021,rudolph_fully_2021,roth_towards_2021,bae_pni_2023,salehi_multiresolution_2021,salehi_multiresolution_2021,wang_student-teacher_2021,deng_anomaly_2022,lee_cfa_2022,thomine_mixedteacher_2023} with several approaches such as normalizing flow\cite{rezende_variational_2016}, memory bank and knowledge distillation \cite{hinton_distilling_2015}.
Our approach is based on a knowledge distillation approach, primarily due to its advantages in terms of inference speed and its well-established capability in detecting defects on various types of textures \cite{thomine_mixedteacher_2023}.
Within the framework of knowledge distillation, knowledge transfer occurs between a teacher-student pair during the training. In the context of unsupervised anomaly detection, where the student model is trained solely on normal samples, the student model generates disparate representations compared to the teacher model when confronted with an anomalous sample. Our approach draws inspiration from the reverse distillation architecture \cite{deng_anomaly_2022}. This architecture follows a student-teacher paradigm, where the teacher model serves as the encoder and the student model as the decoder. By employing this architecture, we aim to leverage the knowledge and guidance provided by the teacher model to enhance the reconstruction capabilities of the student model in the context of anomaly detection. By design, this architecture emphasizes the deeper selected layers and therefore neglects the shallower reconstructed layers. In the domain of texture defect detection, the utmost significance lies in the high-level attributes, as the contextual information has less significance \cite{thomine_mixedteacher_2023}. 
Since we aim at detecting defects on fabrics-like textures, we want to keep shallower information and to that extent we introduced residual connections between the teacher and the student along with custom attention layers. Another crucial component of the reverse distillation approach is the embedding bottleneck, which is responsible for gathering features from specific layers of the teacher model and creating a feature map that serves as input for the student model. In our bottleneck embedding architecture, we strive to design a texture-specific approach that selects the most representative features from the teacher model without altering the descriptive capabilities.
To that extent, we carefully designed a bottleneck based mainly on 1×1 convolutions along with a custom attention block and an SSPCAB layer \cite{ristea_self-supervised_2022}. 
In industrial deployment, interrupting the production chain for training a model on a specific type of texture is not feasible. Hence, we propose a slight modification of our model to introduce a domain-generalized model capable of detecting defects in any plain fabric. The primary idea behind this approach is to enable defect detection while a more precise and specific model is being trained. By adopting this strategy, we can ensure that the production chain remains operational, albeit with a concession in terms of precision. 
Our approach offers outstanding results for defect detection on plain textures, along with a fast inference time and domain generalization capabilities compared to the other state-of-the-art methods.

The primary contributions of this paper are outlined as follows:
\begin{itemize}
  \item  A robust texture anomaly detector utilizing a reverse knowledge-distillation technique. This approach enhances the model ability to detect anomalies in fabric-like textures with improved accuracy and reliability. A slight modification of this approach is also proposed for domain generalization purposes.
  \item Introduction of a novel dataset encompassing a diverse range of fabrics and defects. This dataset provides a valuable resource for training and evaluating fabric anomaly detection models, facilitating research and advancements in the textile domain. 
\end{itemize}

Following the introduction, the subsequent section of this paper is dedicated to reviewing existing literature concerning both feature extraction and deep learning approaches for unsupervised anomaly detection. Subsequently, we present our novel reverse distillation approach with a precise description of the model. The following two sections focus on introducing a new dataset and conducting experiments to evaluate the effectiveness of our method. The final section provides a conclusion for the paper.

\section{Related work}\label{sec2}

To handle the problem of texture defect detection, we first have to extract the main characteristics that describe the texture. As mentioned in \cite{humeau-heurtier_texture_2019}, we can distinguish seven categories of approaches to handle this problem namely statistical approaches, structural approaches, transform-based approaches, model-based approaches, graph-based approaches, learning-based approaches and entropy-based approaches. Each of these approaches offers different advantages and drawbacks and some of them offer competitive results while being really fast with recent computing ressources such as GLCM \cite{partio_rock_nodate} and local binary pattern \cite{ojala_multiresolution_2002} \cite{chen_texture_2005}.\\
More recently, deep learning based feature extraction brought outstanding results for extracting the most important features characterizing a texture. An example of supervised training approaches consists of training a deep texture classifier and then extracting the last layer to obtain informative features. \cite{simon_deep_2020} \\
Autoencoder-based approaches are the most common unsupervised approaches to extract features \cite{kramer_nonlinear_1991} but still suffer from high generalization capabilities which can prevent accurate defect detection.

To tackle the problem of defect detection in the industrial case, gathering every defect of a given object or texture is a laborious task and can lead to poor performances if every type of defect was not considered\cite{han_adbench_2022}. 
Certain methods still utilize supervised training methods like mobile-unet \cite{jing_mobile-unet_2020} which integrates classification and segmentation techniques, or the utilization of SSD-based approaches \cite{xie_improved_2021}. \
Unsupervised anomaly detection deals with the problem of detecting defects on an object or a texture without any prior on the type of possible defects. 
Consequently, many methods emerged proposing different types of algorithms such as autoencoders  
\cite{mei_automatic_2018} and  variational autoencoder variants \cite{nguyen_gee_2019} \cite{zavrtanik_draem_2021}. Another common way of detecting anomalies is Generative Adversarial Networks (GAN),introduced by \cite{goodfellow_generative_2014}, adapted to unsupervised anomaly detection such as Ano-GAN \cite{schlegl_f-anogan_2019}, G2D \cite{pourreza_g2d_2021} and OCR-GAN \cite{liang_omni-frequency_2022}.
More recently, approaches using a classifier pretrained on natural images as feature extractor made their way in industrial anomaly detection, offering outstanding performance. These methods, such as normalizing flow, knowledge distillation and memory banks, use selected layers output features to obtain descriptive feature maps. Then, a training is conducted based on the prior that this feature maps describe the most important characteristics of the given object. 
The normalizing flow approach consists of a flow training based on relevant features of good samples from a pretrained network such as AlexNet \cite{he_deep_2015}, Resnet \cite{krizhevsky_imagenet_2017}, efficient-net \cite{tan_efficientnet_2020} and even vision transformers \cite{vaswani_attention_2017} \cite{dosovitskiy_image_2021} trained on imageNet. Different strategies were used to enhance performances, such as a 2D flow \cite{yu_fastflow_2021} or a cross-scale flow \cite{rudolph_fully_2021}. The use of a memory bank has the advantage of offering a description over the whole class to compare and detect if there is an anomaly \cite{roth_towards_2021} \cite{bae_pni_2023}. \\
Distilling the knowledge from a teacher to a student \cite{hinton_distilling_2015} was recently adapted for unsupervised anomaly detection \cite{salehi_multiresolution_2021} \cite{wang_student-teacher_2021}.
This concept involves training a student network on good samples using the output features of a teacher network that has already been pre-trained for classification purposes. In the test phase, the student network will be capable of replicating the output features of the teacher network when provided with good samples, but when confronted with anomalous samples, the student will not be able to extract descriptive features and therefore a relevant anomaly score can be computed.

Several methods used this principle with different strategies such as a multi-layer feature selection \cite{wang_student-teacher_2021}, a reverse distillation approach \cite{deng_anomaly_2022}, a coupled-hypersphere-based feature adaptation \cite{lee_cfa_2022} and a mixed-teacher approach \cite{thomine_mixedteacher_2023}.

\section{Residual Reverse distillation using specifically selected embeddings}\label{sec3}
This part is dedicated to the use of reverse distillation to detect defects in textures. To this extent, we divided this chapter in four subsections: the reverse distillation approach along with the residual connections, the description of the bottleneck embedding that transforms the teacher's output into input for the student model, the loss and score calculation strategy and the adaptation for domain generalization.

\subsection{Residual reverse distillation}
\textbf{Approach}: The idea of reverse distillation was introduced in \cite{deng_anomaly_2022} with the purpose of proposing an encoder-decoder version of the knowledge distillation paradigm applied to unsupervised anomaly detection. An important problem of unsupervised anomaly detection is the generalization capability, which is problematic since our purpose is to be specific to a texture for detecting any kind of defect. To tackle this specific problem, we propose a residual reverse distillation with a reconstructed embedding emphasizing high-level features. The main idea is to extract the most relevant features describing the texture using a limited encoding and residual connections, allowing a selective and descriptive embedding for input to the decoder student.

\noindent \textbf{Problem definition}: Given a training dataset of images without anomaly ${D=[I_1,I_2,...,I_n]}$, our goal is to extract the information of the $L$ bottom layers. For an image ${I_k} \in R^{w*h*c}$ where $w$ is the width, $h$ the height, $c$ the number of channels and l the l$th$ bottom layer, the teacher outputs features $F_t^l(I_k) \in R^{w_l*h_l*c_l}$ . The decoder output features  are defined as $F_s^{L-l}(I_k) \in R^{w_l*h_l*c_l}$. During the training phase, the student model is trained to reproduce the teacher features on good samples. In testing, when inferring an anomaly, the teacher and the student will output significantly different features and thus will give us an anomaly score.  In the anomaly detection setting,
normal samples follow the same distribution in both $F_t$ and $F_s$. Out-of-distribution samples are considered anomalies.

Our reverse distillation backbone draws inspiration from the one outlined in reverse distillation \cite{deng_anomaly_2022}, differing in terms of layer selection, selective embedding, and residual connections. Given the texture specific architecture described in \cite{thomine_mixedteacher_2023}, we have decided to put an emphasis on the first layers of the network, since they describe high-level features that are primordial in patterned textures.
To this purpose, we employed adaptive loss functions to assign varying significance to individual layers, not to mention that with identical loss factors for every layer, we determine that the loss is more impacted by deeper layers with random weight initialization in the student. Also, we include the initial ResNet layer before the first residual block to guarantee a high-level representation of the texture.  \\
The goal of the teacher encoder T in the reverse distillation paradigm is to extract thorough representations so that the decoder student S can reconstruct the different selected teacher's features. As in previous work \cite{deng_anomaly_2022} \cite{lee_cfa_2022}, we used ResNet pretrained on ImageNet as our backbone for T. As in any knowledge distillation training, the teacher T is frozen to avoid converging to a trivial solution for T and S. In \cite{deng_anomaly_2022}, they used the same exact number of layer for the student and the teacher to match the intermediate representation. We demonstrate that a shallower student with less residual blocks between each dimension is capable of offering a good representation, which also prevents the student to generalize in case of defective sample and therefore increase our defect detection capability in textures. \\

\begin{figure*}[h]
\centerline{\includegraphics[scale=0.3]{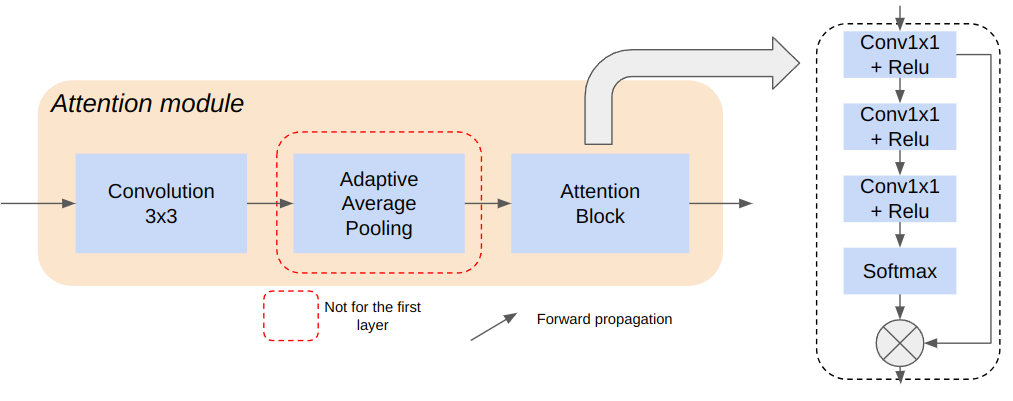}}
\renewcommand{\arraystretch}{1}
\caption{ 
Attention module used for residual connection}
\label{Fig.1}
\end{figure*}

As part of its inherent design, the traditional reverse distillation approach highlights the deeper selected layers, owing to their dimensional proximity with the embedding representation. As mentioned earlier, for textures anomaly detection, shallower layers are more descriptive and less subject to misinterpretation. In this respect, we had some residual connections between the teacher and the student to try to compensate of the loss of information caused by the embedding. However, adding a direct connection between the equivalent layers of the student and the teacher would most likely result in identity or zero-weights depending on the selected method of connection. To counteract this issue, we choose to connect the teacher layer $l$ to the student layer $L-l-1$ with a convolution layer and a pooling layer as well as a self-attention layer, shown in figure \ref{Fig.1}.

With the purpose of enriching the feature representation and modulating the feature importance, we introduce a custom attention block, described in figure 1, which is similar to a spatial attention but uniquely relies on 1×1 convolutions. The inclusion of this layer enables the introduction of non-linearity, leading to improved model performance with minimal additional computational expense.

\begin{figure*}[h]
\centerline{\includegraphics[scale=0.3]{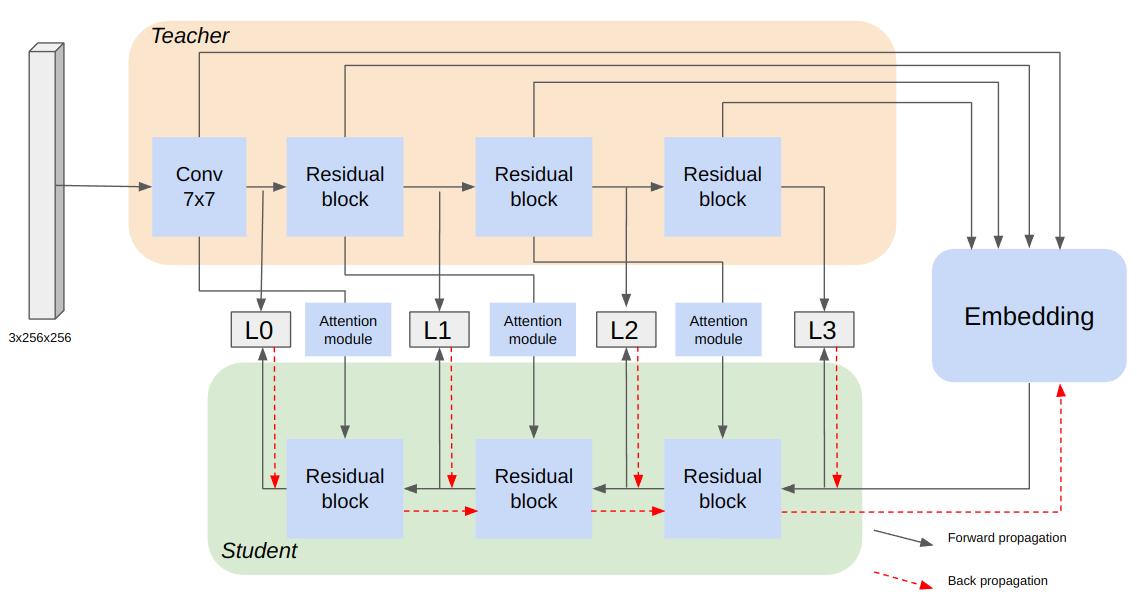}}
\renewcommand{\arraystretch}{1}
\caption{ 
Reverse residual distillation architecture. }
\label{Fig.2}
\end{figure*}

The complete architecture is shown in figure \ref{Fig.2}

\subsection{Bottleneck Embedding}

\begin{figure*}[h]
\centerline{\includegraphics[scale=0.4]{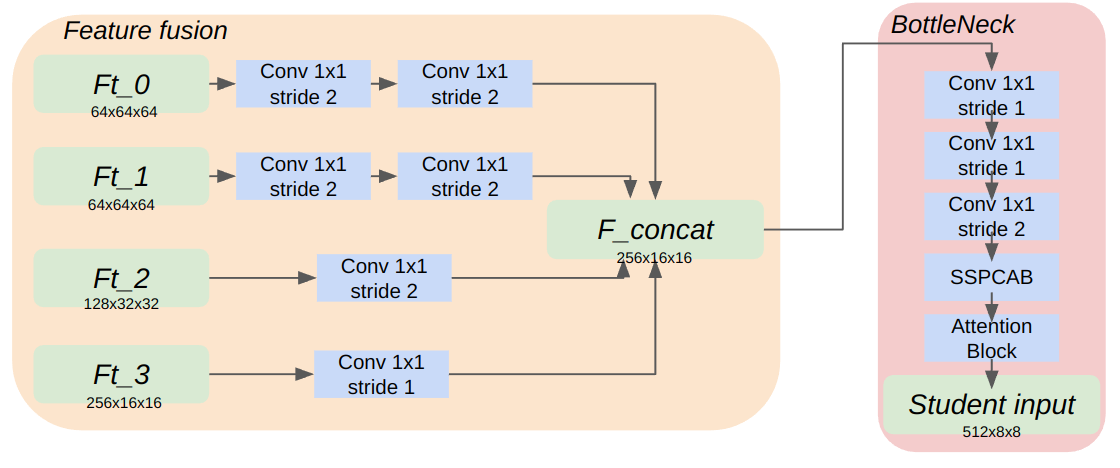}}
\renewcommand{\arraystretch}{1}
\caption{ 
Feature fusion and bottleneck architecture}
\label{Fig.3}
\end{figure*}

One of the main contributions of the reverse distillation paradigm is the embedding representation. In other knowledge-distillation anomaly detector models, the main drawback was the student capacity which could eventually generalize to anomalous defects. MixedTeacher \cite{thomine_mixedteacher_2023} proposed a smaller student to counteract this problem and manage to increase anomaly detection on textures by a huge factor compared to the original STPM model \cite{wang_student-teacher_2021}. 
Our approach goes even further with the selective embedding encoding approach that prevents the student from any unnecessary parameters. 
This approach is specifically outlined for texture defect detection; however, for object defect detection, the size of the embedded minimal representation could potentially vary. This issue revolves around striking a balance between the model's capacity and its capability to accurately reconstruct correct elements, all the while ensuring it remains incapable of reconstructing defective elements.\\ \\
First, to aggregate the features from the selected teacher's layers, we used convolution of size 1 kernel as described in figure \ref{Fig.3} to get all the features at the size of the last layer divided by 2 to prevent the prevalence of the last layer over the others. The obtained feature map is our bottleneck input and the bottleneck embedding will then learn to reconstruct this representation before giving to the student decoder.\\
To increase model performance, we have decided to include a SSPCAB layer \cite{ristea_self-supervised_2022} as the first layer of our bottleneck network since this specific type of layer shows increased performance for a wide variety of unsupervised anomaly detector. To make sure the local information was preserved, we decided to keep a relatively shallow encoding embedding. Each layer is constituted of a convolutionnal layer with a kernel size of 1 and a stride of 1 followed by a ReLu and a BatchNorm layer \cite{ioffe_batch_2015} except for the last which have a stride of 2 to fit the decoder input. Employing convolutional operations with a filter size of 1 confers the benefit of computational efficiency, alongside the capacity to introduce non-linearities and trainable parameters.

\subsection{Loss and Anomaly Score Calculation}
Mathematically, given the problem definition parameter, the pixel-loss function along the channel axis is defined as: \\
\begin{equation}
M^{l}(I_k)_{ij}=\frac{1}{2}\lVert norm(F_t^l(I_k)_{ij})-norm(F_s^{L-l}(I_k)_{ij}) \rVert 
\label{eq.1}
\end{equation}
with $M^l \in \mathbb{R}^{H_l*W_l}$,the layer l loss function as :
\begin{equation}
loss^{l}(I_k)=\frac{1}{w_lh_l}  \sum_{i=1}^{w_l} \sum_{j=1}^{h_l} M^l(I_k)_{ij}  
\label{eq.2}
\end{equation}

\noindent  and the global loss is written as: 

\begin{equation}
loss(I_k)= \sum^{l} \alpha_l loss^{l}(I_k) 
\label{eq.3}
\end{equation}
with $\alpha_l$ the loss factor for the layer l.

At the testing stage, we want an anomaly score and an anomaly localization. To begin with, let's address the anomaly localization problem. The paradigm of knowledge distillation for anomaly detection and localization stipulates that, in case of anomalous sample, the teacher is able to represent the defect  while the student lacks generalization capability as it has learnt to generate features only for defect-free samples. This means that the student and the teacher features will be different and this will reflect on the anomaly map calculated in equation \ref{eq.1}.
To account for the anomaly maps of all the features, we regroup all these features maps with upsampling in a single localization map denoted $L_{map}$:
\begin{equation}
L_{map} = \sum^{l}_{i=1} UpSample(M^l) 
\label{eq.4}
\end{equation}
A Gaussian filter is then used on $L_{map}$ to smooth the defect localization. \\
For the image anomaly score or defect detection, to avoid the prevalence of huge defects over small defects, we decided not to use the mean of the anomaly map but the max value instead based on the paradigm that a non-defective anomaly map will have no significant out-of-distribution value especially after the application of the Gaussian filter. 

\subsection{Reverse distillation for domain generalization}
Domain generalization in deep learning refers to the task of training a model on data from multiple source domains, and then generalizing this model to perform well on a new target domain. 
The goal of domain generalization is to improve the generalization performance of deep learning models, especially in situations where the test data comes from a different distribution than the training data. 
In the specific domain of fabric defect detection, our objective is to develop an anomaly detection system capable of identifying defects in all types of textiles. However, it is important to note that our design and testing procedures are specifically tailored for homogeneous textures and may not be directly applicable to fabric designs that exhibit more intricate patterns or variations. As demonstrated in FABLE \cite{thomine_fable_2023}, the knowledge distillation architecture has outstanding domain-generalized performance for homogeneous textures. \\
The domain generalization approach for fabric anomaly detection can be viewed as a strategy that provides additional training samples to the model. By incorporating a diverse range of fabric samples during training, the model becomes more adept at recognizing fabric-specific characteristics, such as thread patterns and lattice structures. This broader exposure helps the model generalize better across different fabric types, enhancing its ability to detect anomalies across a variety of textile materials. Indeed, while the idea of learning inherent textile patterns is appealing, achieving effective fabric anomaly detection is a complex task. There are various challenges associated with the intricacies of fabric textures, including variations in weave patterns, colors, textures, and defect types. Additionally, factors such as lighting conditions, scale, and resolution can impact the detection accuracy. Developing a robust and accurate fabric anomaly detection model requires careful consideration of these complexities and the implementation of sophisticated techniques to account for them. \\
As previously discussed, simpler patterns like thread and lattices are often captured by the shallower layers of the model architecture. This observation motivated the inclusion of a residual connection, which places emphasis on these particular layers. By doing so, the model can effectively capture and learn these fundamental fabric patterns, thereby enhancing its ability to detect anomalies associated with thread and lattice structures. \\
On the contrary, in the case of domain generalization, we believe that all deeper layers are a key to understanding the intricacies of such a descriptive problem. Consequently, we have made the decision to eliminate the residual connection mentioned in Section 3.1.1, as we consider it to be less relevant in this context.

\section{Industrial textile dataset}
In this part, we introduce the industrial fabrics dataset containing 2150 images with 6 types/colors of textile. We constitute two nomenclature for both domain-generalized and classic unsupervised anomaly detection training. We decide to use the same folder architecture as MVTEC AD dataset \cite{bergmann_mvtec_2019}. \\

\noindent To constitute the dataset, we placed 2 cameras from different brand in front of a visiting machine with strong and nearly uniform light source. \\
\subsection{Industry constraints}
Industrial deployments are subject to certain constraints pertaining to the positioning of cameras and the intensity of light sources to prevent any disruption to the machine operator's activities.
Our aim was to carefully choose an ideal pairing of light intensity and shutter duration to guarantee the capture of images that are sharp and clearly defined.
However, due to variations in fabric color and the velocity of the visiting machine, our chosen light source did not consistently provide adequate brightness.
As a result, certain images within the dataset exhibit a slight degree of blurriness, contributing to the creation of a more demanding and authentic dataset. This reflects the fact that achieving flawless conditions, even in an industrial setting, is frequently unachievable.
\subsection{Variety of defects}
\begin{figure*}[h]
\centerline{\includegraphics[scale=0.2]{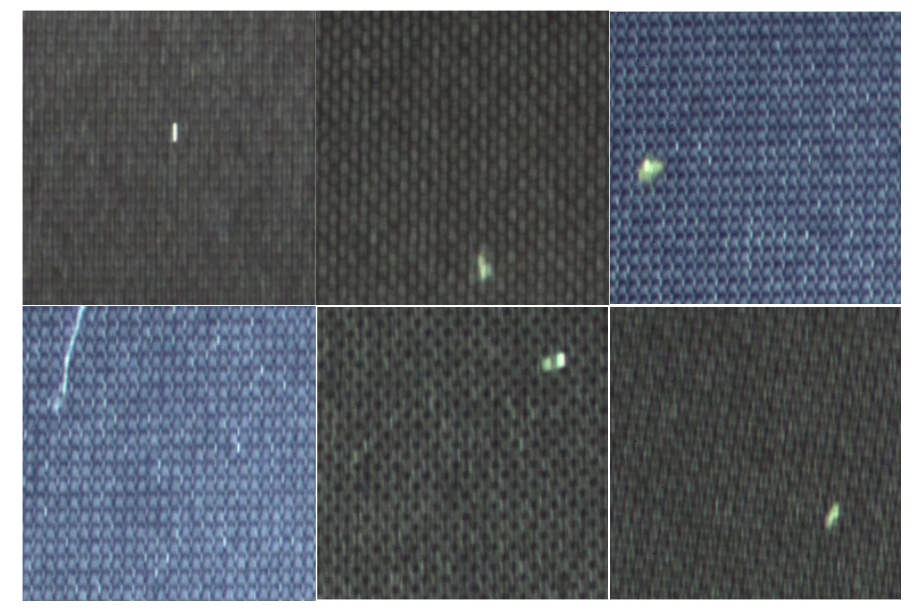}}
\renewcommand{\arraystretch}{1}
\caption{ 
Images extracted from our proposed dataset with defect that could be dust}
\label{Fig.4}
\end{figure*}

As previously stated, defects within the textile industry exhibit a wide range of variations. Furthermore, our dataset will continue to expand as we accumulate additional instances of defects through the ongoing use of our cameras. 
In addition, it is worth noting that certain anomalies present in our database may consist of dust particles that have settled on the textile, which may not typically be classified as defects in most textile industries. However, due to our inability to definitively determine whether these anomalies are genuine defects or simply dust, we have chosen to include them in the dataset. Figure \ref{Fig.4} showcases examples of such anomalies.

\captionsetup{justification=centering}

\subsection{Dataset description}
\begin{figure*}[h]
\centerline{\includegraphics[scale=0.35]{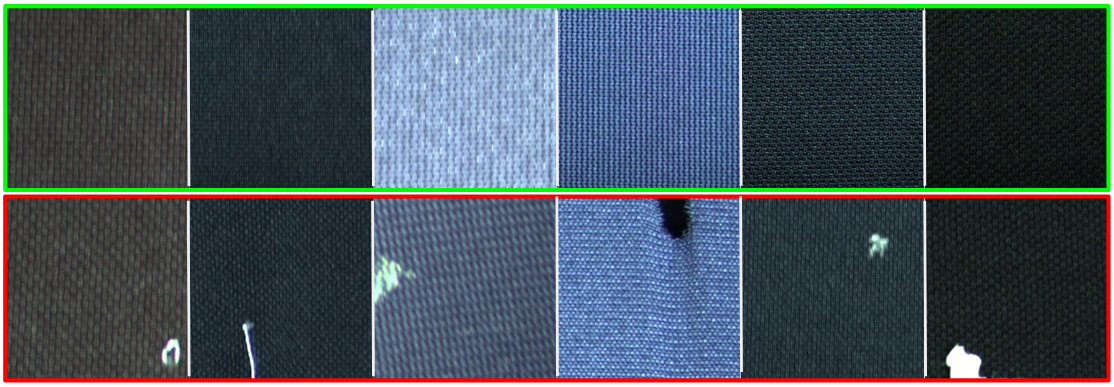}}
\renewcommand{\arraystretch}{1}
\caption{ 
Images of our proposed datasets, (upper) defect-free images, (lower) defective images. Some images are a bit blurry due to motion blur. Other kind of defective samples are presented in the experiment part. }
\label{Fig. dataset}
\end{figure*}

The dataset is composed of patches of size 256x256 extracted from 2 industrial cameras of resolution 3088x2076 and 2464x2056 respectively. The defects included in the dataset are not artificially generated but rather originate from real-world instances within the textile industry.  However, we added some non-defect from the industry like writing on the fabric or the sewing attaching two pieces of fabric. These added "defects" aim to diversify the dataset and provide a broader range of defect types for training and analysis purposes.
The dataset contains 6 classes with between 199 and 588 images for training and 30 and 114 for testing.
The precise distribution is presented in table \ref{Table Def}. 
It is important to acknowledge that the dataset poses a potential long-tail distribution challenge, especially for domain generalization training, but remains suitable for “classic” unsupervised anomaly detection where distinct models are trained for each class.
An overview of the dataset can be seen in figure \ref{Fig. dataset}.

\captionsetup{justification=centering}

\begin{table}
\begin{tabular}{|c|cccc|} 
 \hline
		 \textbf{Category} & \textbf{total} & \textbf{train}& \textbf{testGood}&
        \textbf{testDefective} \\
        \hline
		{type1cam1} &
		386  & 272 & 28  & 86 \\
			
    	{type2cam2} & 
		257  & 199  & 19 & 39 \\
		
		{type3cam1} & 
		689  & 588  & 54 & 47 \\
		
		{type4cam2} &
		  229  & 199  & 19 & 11 \\

        {type5cam2} &
		  298  & 199  & 19 & 80 \\

        {type6cam2} &
		  291  & 199  & 19 & 73 \\
		
		\hline
		{total} &
		2150  & 1656 &  158 & 336 \\
		
	    \hline
\end{tabular}
\caption{Industrial textile dataset distribution}
\label{Table Def}
\end{table}

\subsection{Existing datasets}
In order to provide contextualization and to position our dataset within the broader landscape of existing datasets, it is essential to present other relevant datasets that specifically address the problem of texture anomaly detection. \\
One of the widely used industrial unsupervised anomaly detection datasets is the MVTEC AD dataset\cite{bergmann_mvtec_2019}, which comprises 10 objects and 5 textures. While 4 of these textures align well with the problem of textile anomaly detection, one texture is significantly different. 
Other datasets, such as TILDA \cite{deutsche_forschungsgemeinschaft_tilda_nodate}, Aitex \cite{silvestre-blanes_public_2019} and DAGM \cite{matthias_wieler_tobias_hahn_fred_a_hamprecht_weakly_2007} , offer grayscale fabrics-like datasets for anomaly detection purposes. In addition, datasets like BTech \cite{mishra_vt-adl_2021}, NEU-CLS \cite{bao_triplet-graph_2021}, KolektorSDD \cite{tabernik_segmentation-based_2019}, Magnetic Tile Dataset \cite{huang_y_qiu_c_guo_y_wang_x__yuan_k_surface_2018} and RSDDs \cite{lu_scueu-net_2020} provide supplementary datasets for surface defect detection, which are not specific to textiles. \\
Our dataset strives to offer color images of fabrics captured under authentic conditions, showcasing a wide range of real defects. With an emphasis on diversity, our dataset encompasses an extensive array of defect types, ensuring its applicability in addressing various anomaly detection challenges within the textile industry.

\section{Experiments}
This part showcases performances of our knowledge distillation approach both in terms of AUROC score and inference time since these are the most concerning application requirements in anomaly detection. \\
We have also added a coverage measure to determine the actual benefit of this approach in real case scenario. \\
Training and inference were done on an RTX 3080ti.
To ensure efficient inference for industrial applications, we employ ResNet18 and ResNet34 as the backbone architectures for our anomaly detection system. The teacher model, which serves as the reference model, is a ResNet34 trained on the ImageNet dataset. In contrast, the student model consists of a ResNet18 backbone. It is worth noting that the student model is structured in a backward manner, as explained in the part 3.
 We used ADAM optimizer\cite{kingma_adam_2017}, a learning rate of 0.005 with scheduler to reduce the learning rate when hitting a plateau during the training. The training was done for 100 epochs with a batch size of 4. 
We resized all the images to a size of 256x256 keeping respectively 70\% and 30\% for training and validation. We kept the checkpoint with the lowest validation loss and introduce early stopping.

\subsection{State of the art AUROC comparison}
To evaluate our architecture, we use several datasets including MVTEC AD textures \cite{bergmann_mvtec_2019}, TILDA \cite{deutsche_forschungsgemeinschaft_tilda_nodate} dataset, AITEX dataset \cite{silvestre-blanes_public_2019} and our dataset presented in section 4. \\
For MVTEC AD, we selected the four fabric-like textures and compared our results to the other state-of-the-art methods. The results are presented in table \ref{Table 1} for anomaly detection and table \ref{Table 2} for anomaly localization.\\

\begin{table}
\begin{tabular}{|c|ccccc|c|} 
 \hline
		 \textbf{Category} & \textbf{CFA} & \textbf{PatchCore\cite{roth_towards_2021}}& \textbf{FastFlow\cite{yu_fastflow_2021}}&
        \textbf{Reverse distillation\cite{deng_anomaly_2022}} & \textbf{MixedTeacher\cite{thomine_mixedteacher_2023}} & \textbf{Ours} \\
        \hline
		{carpet} &
		97.3  & 98.7 & 99.4 & 98.9 & 99.8 & \textbf{100} \\
			
    	{tile} &
		99.4  & 98.7 & 100 & 99.3 & \textbf{100} & 99.7 \\
		
		{wood} &
		99.7  & 99.2 & 99.2 & 99.2 & 99.6 & \textbf{99.8} \\
		
		{leather} &
		  100  & 100 & 99.9 & 100 & 100 & \textbf{100} \\
		
		\hline
		{Mean} &
		99.1  & 99.1 & 99.6 & 99.3 & 99.8 & \textbf{99.9} \\
		
	    \hline
\end{tabular}
\caption{Anomaly detection results with AUROC on MVTEC fabric-like textures}
\label{Table 1}
\end{table}

\begin{table}
\begin{tabular}{|c|cccc|c|} 
 \hline
		 \textbf{Category} & \textbf{PatchCore\cite{roth_towards_2021}}& \textbf{FastFlow\cite{yu_fastflow_2021}}&
        \textbf{Reverse distillation\cite{deng_anomaly_2022}} & \textbf{MixedTeacher\cite{thomine_mixedteacher_2023}} & \textbf{Ours} \\
        \hline
		{carpet} &
		98.9   & \textbf{99.1} & 98.9 & 99.0 & 98.8 \\
			
    	{tile} &
		95.6   & \textbf{96.6} & 95.6 & 95.9 & 93.7 \\
		
		{wood} &
		95   & 94.1 & \textbf{95.3} & 94.9 & 92.4 \\
		
		{leather} &
		  99.3   & \textbf{99.6} & 99.4 & 99.6 & 98.7 \\
		
		\hline
		{Mean} &
		97.2   & \textbf{97.3} & 97.3 & 97.4 & 95.9 \\
		
	    \hline
\end{tabular}
\caption{Anomaly localization results with AUROC on MVTEC fabric-like textures}
\label{Table 2}
\end{table}

\captionsetup{justification=centering}
\begin{figure*}[h]
\centerline{\includegraphics[scale=0.3]{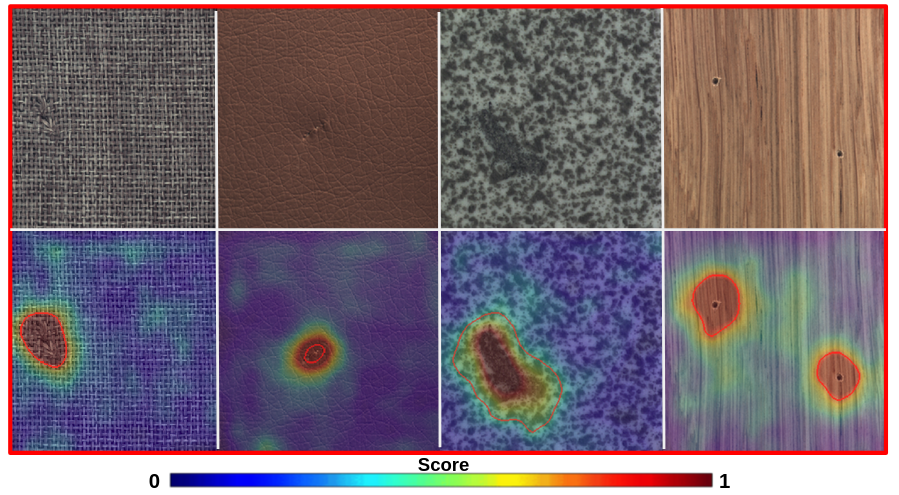}}
\renewcommand{\arraystretch}{1}
\caption{ 
Input image and anomaly map in the 4 fabric-like MVTEC AD textures}

\label{Fig.8}
\end{figure*}

\begin{figure*}[h]
\centerline{\includegraphics[scale=0.3]{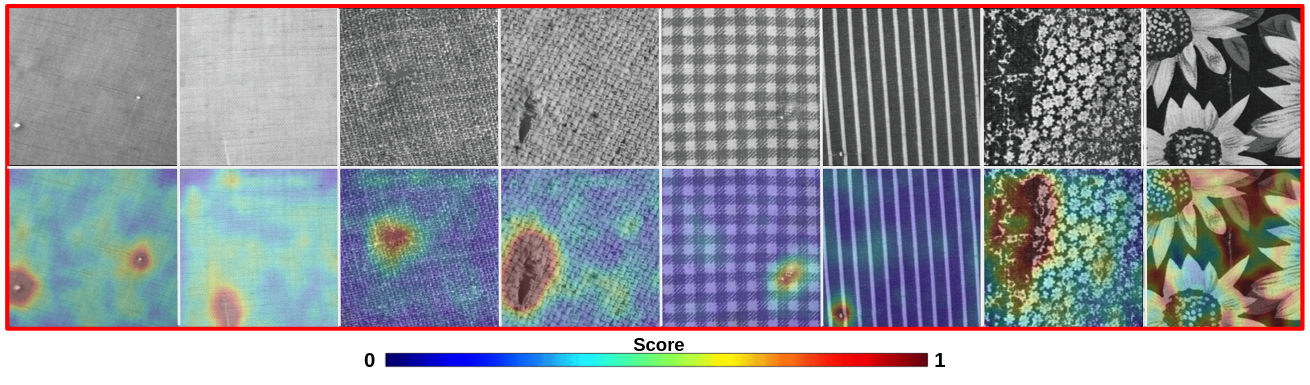}}
\renewcommand{\arraystretch}{1}
\caption{ 
Input image and anomaly map in all 8 TILDA textile. As expected, the model is too restrictive to handle textures with patterns like flowers}

\label{Fig.5}
\end{figure*}

For the TILDA dataset, we restructured the data to conform to the MVTEC nomenclature. To ensure a fair comparison with state-of-the-art methods, the anomalib library \cite{akcay_anomalib_2022} was employed.
The results are shown in table \ref{Table 4}. 
Examples of input images and results can be seen in figure \ref{Fig.5}\\

\begin{table}
\resizebox{15cm}{!}{
\begin{tabular}{|c|cccccccc|c|} 
 \hline
		 \textbf{Tilda} & \textbf{fabric1} & \textbf{fabric2}& \textbf{fabric3}&
        \textbf{fabric4} & \textbf{fabric5} & \textbf{fabric6} & \textbf{fabric7}
        & \textbf{fabric8} & \textbf{mean}\\
        \hline
        {reverseDistillation\cite{deng_anomaly_2022}} &
		94.8  & 88.2 & 91.4 & 59.6 & 67.4 &  78.7 & \textbf{78.6} & \textbf{84.5} & 80.4 \\
		{ours} &
		\textbf{96.9}  & \textbf{95.8} & \textbf{92.5} & \textbf{75.0} & \textbf{87.2} &  \textbf{88.6} & 70.7 & 61.9 & \textbf{83.6} \\

	    \hline
\end{tabular}
}
\caption{Anomaly detection results with AUROC on Tilda textures}

\label{Table 4}
\end{table}

\captionsetup{justification=centering}

\begin{figure*}[h]
\centerline{\includegraphics[scale=0.3]{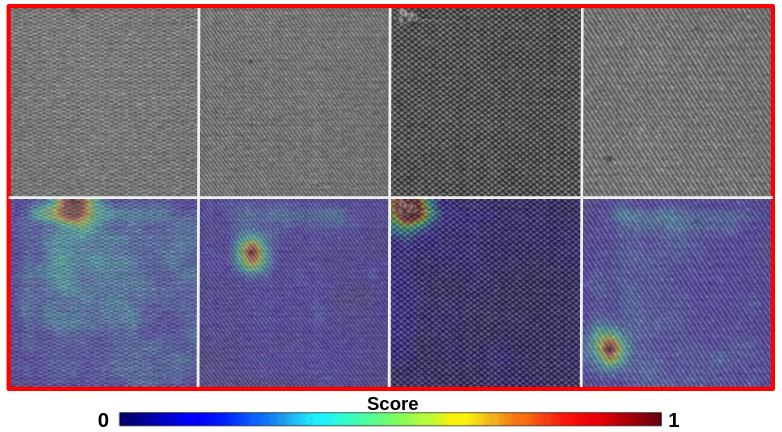}}
\renewcommand{\arraystretch}{1}
\caption{ 
Input image and anomaly map in all 4 Aitex textile}

\label{Fig.6}
\end{figure*}

For the aitex dataset, we separate each 4096x256 images into 16 patches of size 256x256 and used the MVTEC AD nomenclature. The results of our experiment are presented in table \ref{Table 5}. Examples of input images and their corresponding results are visually depicted in figure \ref{Fig.6}

\begin{table}
\begin{tabular}{|c|cccc|c|} 
 \hline
		 \textbf{Aitex} & \textbf{fabric1} & \textbf{fabric2}& \textbf{fabric3}&
        \textbf{fabric4} & \textbf{mean}\\
        \hline
        {reverseDistillation \cite{deng_anomaly_2022}} &
		92.7  & 96.5 & 77.3  & 97.7 & 91.0 \\
		{ours} &
		\textbf{97.8}  & \textbf{99.2} & \textbf{89.1}  & \textbf{100} & \textbf{96.5} \\

	    \hline
\end{tabular}
\caption{Anomaly detection results with AUROC on Aitex textures}
\label{Table 5}
\end{table}

\captionsetup{justification=centering}

\begin{figure*}[h]
\centerline{\includegraphics[scale=0.3]{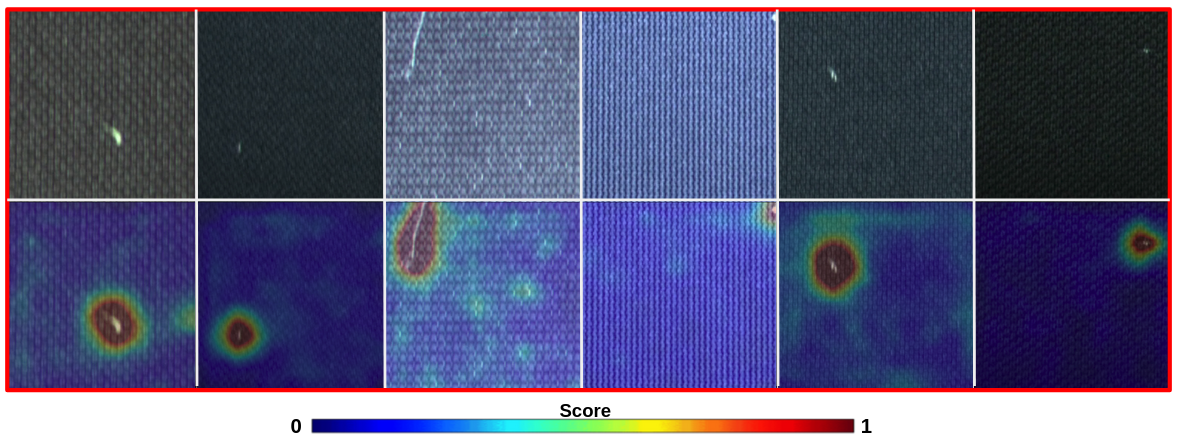}}
\renewcommand{\arraystretch}{1}
\caption{ 
Input image and anomaly map in industrial textile database. We showcase speed-blurry image on purpose to demonstrate our capability to deal with such quality differences}

\label{Fig.ITDB}
\end{figure*}

Subsequently, we conduct assessments on the dataset for which our model has been meticulously tailored. This dataset is derived from an authentic industrial process, aligning with our model's primary objective of optimizing performance specifically for this designated task. Results are depicted in table \ref{Table FT} and images and anomaly maps are shown in figure \ref{Fig.ITDB}.
Our results on the dataset are exceptionally accurate as our model was specifically designed and optimized for it. However, we continue to record images using the camera and actively strive to gather a larger and more diverse dataset. By expanding the dataset, we aim to introduce greater challenges and complexities, which will further enhance the robustness and generalization capabilities of our model.
\captionsetup{justification=centering}

\begin{table}
\resizebox{15cm}{!}{
\begin{tabular}{|c|cccccc|c|} 
 \hline
		 \textbf{ITD} & \textbf{type1cam1} & \textbf{type2cam2}& \textbf{type3cam1}&
        \textbf{type4cam2} & \textbf{type5cam2} & \textbf{type6cam2} & \textbf{mean}\\
        \hline
        {reverseDistillation \cite{deng_anomaly_2022}} &
		100  & 100 & 98.3 & 100 & 100 &  100 & 99.7 \\
		{ours} &
		100  & 100 & \textbf{100} & 100 & 100 &  100 & \textbf{100}  \\

	    \hline
\end{tabular}
}
\caption{Anomaly detection results with AUROC on Industrial textile dataset}
\label{Table FT}
\end{table}

The domain generalization results are presented in table \ref{Table 3}
\begin{table}
\begin{tabular}{|c|ccccc|c|} 
 \hline
		 \textbf{Category} & \textbf{Epi-FRC+ \cite{li_episodic_2019}} & \textbf{EISNet+\cite{wang_learning_2020}}& \textbf{DGTSAD \cite{chen_domain-generalized_2022}}&
        \textbf{Fable \cite{thomine_fable_2023}} & \textbf{Ours} \\
        \hline
		{carpet} &
		91.6  & 98.2  & 94.3 & 98.5 & \textbf{99.3} \\
			
    	{tile} &
		95.1  & 85.1  & 99.4 & 96.5 & \textbf{98.8} \\
		
		{wood} &
		94.1  & 97.9  & 96.2 & \textbf{99.9} & 99.7 \\
		
		{leather} &
		  100  & 100  & \textbf{100} & 99.1 & 99.9 \\
		
		\hline
		{Mean} &
		95.2 & 95.3 & 97.5 & 98.5 & \textbf{99.4} \\
		
	    \hline
\end{tabular}
\caption{Domain generalization comparison}
\label{Table 3}
\end{table}

Results interpretation : Results on the TILDA dataset demonstrates that our approach offers outstanding results on plain fabrics and simple patterns such as checkered or striped. The approach is therefore limited for more complex patterns that require more contextual information and a higher receptive field. That consideration establishes the limitation of such a restrictive architecture that emphasizes mostly high-level features along with 1×1 convolutions that do not increase the receptive field. Nonetheless, the results on MVTEC AD, AITEX and our dataset show that our model outperforms state-of-the-art approaches for “simple” textures.

\subsection{Inference speed comparison}
When designing the architecture, one of the primary concerns was the inference speed. Our approach of knowledge distillation offers the advantage of fast inference due to the utilization of ”smaller” networks such as ResNet18 and ResNet34. Despite their reduced size, these networks can still deliver state-of-the-art performance, as demonstrated in the previous part. The efficacy of both the bottleneck embedding component and the residual connection component is significantly reliant upon the utilization of 1×1 convolutions, which exert minimal influence on the overall computational velocity. This combination allows for efficient and accelerated anomaly detection inference while maintaining high detection accuracy. \\ 
The added embedding and residual connections are even lighter and therefore our method offer fast inference time. \\
In our specific scenario, we process high resolution images of size 3000x2000 and therefore, we will have to divide this image in smaller patches of size 256x256. The corresponding number of patches is 96 patches. To report industrial plausible inference time, we processed in batch of size 32 and reported the inference time for the whole image processing (3 batches) as shown in table \ref{Table 6}. 

\begin{table}
\begin{tabular}{|c|cc|} 
 \hline
		 \textbf{Batch size} & \textbf{Inference time} & \textbf{FPS} \\
        \hline
		  16 : time per patch & 0.7ms  & 1450   \\ 
			
    	32 : time per patch & 0.5ms & 2000   \\
		
		whole image (96 patches)   & 45ms & 22 \\

	    \hline
\end{tabular}
\caption{Inference speed}
\label{Table 6}
\end{table}

\subsection{Coverage Measure}

Applications of this nature are primarily designed to assist humans in the task of detecting defects in the visiting machine. By leveraging automated anomaly detection methods, the goal is to provide support and assistance to human operators in their inspection and quality control tasks. These applications aim to improve the efficiency and accuracy of defect detection processes, enabling operators to identify and address anomalies more effectively. 
Consequently, we have made the decision to incorporate a metric that accurately reflects the practical impact of our method on assisting industrial processes. This metric will serve as an objective measure to assess the effectiveness and efficiency of our approach in real-world scenarios. \\ 
The coverage metric quantifies the level of confidence exhibited by the model while maintaining a fixed precision and recall. For example, setting a target of 80 percent coverage with a precision of 95 percent would imply that the model will classify 80 percent of defects with a precision of 95 percent, while external assistance would be required for the remaining 20 percent. \\
In an industrial scenario, the tolerance for false negatives is typically extremely low, indicating that the detection of all defects is essential. Therefore, a recall of 100 percent, ensuring that no defects are missed, is a critical requirement for fabric defect detection. Additionally, to prioritize precision and avoid false positives, we have decided to set the precision at 100 percent as well. This means that all identified defects are guaranteed to be genuine, providing a high level of confidence in the accuracy of the defect detection system. In conjunction with the outcomes related to inference speed and domain generalization performance, these results collectively support the potential applicability of our approach within a real-world industrial setting, potentially involving collaboration with a human operator.\\
The table \ref{Table 7} showcases coverage value for the tested MVTEC textures.

\begin{table}
\begin{tabular}{|c|ccc|} 
 \hline
		 \textbf{Category} & \textbf{Coverage}& \textbf{Coverage +/- 0.02} & \textbf{Coverage +/- 0.05}\\
        \hline
		{carpet} &
		100  & 97 & 91   \\
			
    	{tile} &
		91 &  81 & 71  \\
		
		{wood} &
		99  & 96 & 91   \\
		
		{leather} &
		  100  & 100 & 100  \\
		
		\hline
		{Mean} &
		97.5  & 93.5 & 88.3  \\
		
	    \hline
\end{tabular}
\caption{Coverage measure with fixed precision and recall at 100 percent. The +/- indicates the score tolerance to consider that 2 samples are ambiguous. Base coverage is with score tolerance 0}
\label{Table 7}
\end{table}

\section{Conclusion}
In summary, we have introduced a novel unsupervised anomaly detection approach utilizing knowledge distillation principles, with a specific emphasis on reverse distillation. By using thoughtful layer selection and incorporating residual connections, we have achieved exceptional outcomes on the MVTEC anomaly detection dataset, surpassing the performance of previous state-of-the-art methods specifically for these targeted classes. In order to address the challenge of fabric anomaly detection, we conducted extensive testing of our model using a diverse fabric defect database. Our findings validate the superior performance of our unsupervised fabric defect detection approach.
However, the architectural design of our approach is specifically tailored for the problem of plain fabric defect detection and is limited when dealing with big patterns such as drawings.
Due to the lack of existing databases tailored to our specific fabric anomaly detection problem, which involves high-resolution color images captured by industrial cameras in real-world scenarios, we have taken the initiative to create a novel dataset. It comprises 2150 images of diverse fabric types, encompassing several commonly encountered defect types prevalent in the fabric industry. 
Presently, this dataset is limited in terms of the pool of defects gathered since in a fabric factory some defects are rarer than others.
It will, therefore, undergo regular updates to incorporate additional classes and to introduce new types of defects. This continuous expansion aims to enhance the dataset diversity and ensure its relevance to evolving fabric anomaly detection research and industry requirements. This dataset will be freely available for the scientific community.(https://github.com/SimonThomine/IndustrialTextileDataset)

\bibliography{sn-article}

\end{document}